\documentclass[10pt,twocolumn,letterpaper]{article}

\usepackage{cvpr}              % To produce the CAMERA-READY version

\usepackage[accsupp]{axessibility}
\usepackage{graphicx}
\usepackage{amsmath}
\usepackage{amssymb}
\usepackage{booktabs}
\usepackage{algorithm}
\usepackage{algorithmic}
\usepackage [switch]{lineno}
\usepackage{amsfonts}
\usepackage{amsbsy}
\usepackage{amsmath}
\usepackage{booktabs}
\usepackage{multirow}
\usepackage{epsfig}
\usepackage{newfloat}
\usepackage{listings}
\usepackage{bm}

\usepackage[pagebackref,breaklinks,colorlinks]{hyperref}
\makeatletter
\def\blfootnote{\xdef\@thefnmark{}\@footnotetext}
\makeatother

\usepackage[capitalize]{cleveref}
\crefname{section}{Sec.}{Secs.}
\Crefname{section}{Section}{Sections}
\Crefname{table}{Table}{Tables}
\crefname{table}{Tab.}{Tabs.}

\def\cvprPaperID{106} % *** Enter the CVPR Paper ID here
\def\confName{CVPR}
\def\confYear{2023}

\begin{document}

%%%%%%%%% TITLE - PLEASE UPDATE
\title{Instance-Aware Domain Generalization for Face Anti-Spoofing}

\author{Qianyu Zhou$^{1,3}$\footnotemark[1], 
Ke-Yue Zhang$^{2}$\thanks{\textit{Equal contribution. }}, 
Taiping Yao$^2$, 
Xuequan Lu$^4$,
Ran Yi$^{1}$, 
\\Shouhong Ding$^2$\footnotemark[2],
Lizhuang Ma$^1$\thanks{\textit{Corresponding author.}}
\vspace{1mm}
\\$^1$Shanghai Jiao Tong University; $^2$Youtu Lab, Tencent; \\$^3$ Shanghai Key Laboratory of Computer Software Evaluating and Testing; $^4$
Deakin University.\\
$^1${\tt\small \{zhouqianyu,ranyi\}@sjtu.edu.cn}, $^1${\tt\small ma-lz@cs.sjtu.edu.cn},  \\ $^2${\tt\small
\{zkyezhang,taipingyao,ericshding\}@tencent.com}, $^4${\tt\small xuequan.lu@deakin.edu.au }\\
% $^4${\tt\small xuequan.lu@deakin.edu.au} \\
}

\maketitle

%%%%%%%%% ABSTRACT
\begin{abstract}

Face anti-spoofing (FAS) based on domain generalization (DG) has been recently studied  to improve the generalization on unseen scenarios. 
Previous methods typically rely on domain labels to align the distribution of each domain for learning domain-invariant representations. 
However, artificial domain labels are coarse-grained and subjective, which cannot reflect real domain distributions accurately.
Besides, such domain-aware methods focus on domain-level alignment, which is not fine-grained enough to ensure that learned representations are insensitive to domain styles. To address these issues, we propose a novel perspective for DG FAS that aligns features on the instance level without the need for domain labels. Specifically, Instance-Aware Domain Generalization framework is proposed to learn the generalizable feature by weakening the features' sensitivity to instance-specific styles. Concretely, we propose Asymmetric Instance Adaptive Whitening to adaptively eliminate the style-sensitive feature correlation, boosting the generalization. Moreover, Dynamic Kernel Generator and Categorical Style Assembly  are proposed to first extract the instance-specific features and then generate the style-diversified features with large style shifts, respectively, further facilitating the learning of style-insensitive features. Extensive experiments and analysis demonstrate the superiority of our method over state-of-the-art competitors. Code will be publicly available at this  \href{https://github.com/qianyuzqy/IADG}{\textit{link}}.
\end{abstract}

\section{Introduction}

\begin{figure}
\includegraphics[width=\linewidth]{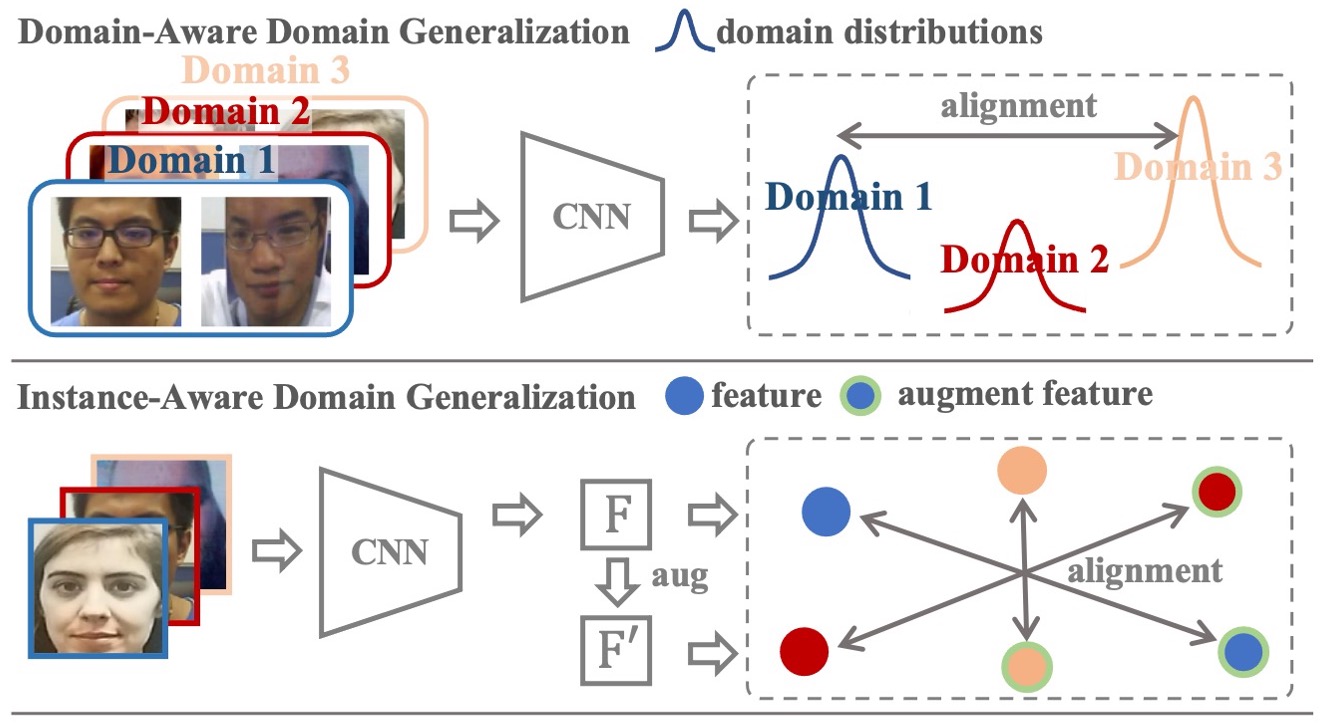}
% \vspace{-5mm}
\caption{Conventional DG-based FAS approaches typically rely on artificially-defined domain labels to perform \textit{domain-aware domain generalization}, which cannot guarantee that the learned representations are still insensitive to domain-specific styles.
In contrast, our method does not rely on such domain labels and focuses on the \textit{instance-aware domain generalization} via exploring asymmetric instance adaptive whiting on the fine-grained instance level.}
\vspace{-4mm}
\label{fig:illustrator}
\end{figure}

Face anti-spoofing (FAS) plays a critical role in protecting face recognition systems from various presentation attacks, \emph{e.g.,} printed photos, video replay, \emph{etc}. 
To cope with these presentation attacks,
a series of FAS works based on hand-crafted features~\cite{boulkenafet2015face,maatta2011face,LBP01,2014Context,HoG01}, and deeply-learned features~\cite{DeepBinary00,yu2021revisiting,zhang2021structure,lin2019face,hu2022structure} have been proposed. Although these methods have achieved promising performance in intra-dataset scenarios, they suffer from poor generalization when adapting to various unseen domains.

To improve the generalization ability on unseen domains, 
recent studies introduce domain generalization (DG) techniques into the FAS tasks, which utilize the adversarial learning~\cite{2019Multi,2020Single,wang2022domain} or meta-learning~\cite{liu2021dual,liu2021adaptive,chen2021generalizable,zhou2022adaptive,du2022energy} to learn domain-invariant representations. %for generalizing on the unseen domains. 
Despite its gratifying progress, most of these DG-based FAS methods utilize domain labels to align the distribution of each domain for domain-invariant representations, as shown in Figure \ref{fig:illustrator}.
However, such domain-aware methods suffer from two major limitations.
Firstly, the artificial domain labels utilized in their methods are very coarse, and cannot accurately and comprehensively reflect the real domain distributions.
For example, numerous illumination conditions, attack types, and background scenes are ignored in the source domains, which might lead to various fine-grained sub-domains. 
Though D$^2$AM~\cite{chen2021generalizable} tries to alleviate these issues via assigning pseudo domain labels to divide the mixed source domains, it still manually sets the number of pseudo source domains and does not solve the problem in essence. 
Secondly, such domain-level alignment only constrains features from the perspective of distribution, which is not fine-grained enough to guarantee that all channels of features are insensitive to the instance-specific styles. 
Thus, the learned features might still contain information sensitive to instance-specific styles when encountering novel samples, failing to generalize on the unseen domain.

To address these issues, we propose a novel perspective of DG-FAS that explores the style-insensitive features and aligns them on a fine-grained instance level without the need for domain labels, improving the generalization abilities towards unseen domains. 
Specifically, we propose an \textit{Instance-Aware Domain Generalization} (IADG) framework to dynamically extract generalized representations for each sample by encouraging their features to be insensitive to the instance-specific styles. 
% Our IADG framework consists of three key components. 
Concretely, we first introduce Asymmetric Instance Adaptive Whitening (AIAW) to boost the generalization of features via adaptively whitening the style-sensitive feature correlation for each instance.
Instead of directly learning the domain-agnostic features, AIAW aims to weaken the feature correlation (\emph{i.e.,} covariance matrix) from higher-order statistics on a fine-grained instance level. 
Considering the distribution discrepancies of real and spoof samples, AIAW adopts asymmetric strategies to supervise them, boosting the generalization capability. 
Moreover, to facilitate the learning of style-insensitive features in AIAW,  Dynamic Kernel Generator (DKG) and Categorical Style Assembly (CSA) are proposed to generate style-diversified features for further AIAW. 
Specifically, DKG models the instance-adaptive features, which automatically generates instance-adaptive filters that work with static filters to facilitate comprehensive instance-aware feature learning.
Based on such instance-adaptive features, CSA simulates instance-wise domain shifts by considering the instance diversity to generate style-diversified samples in a wider feature space, which augments real and spoof faces separately to prevent the label changes in the FAS task.  Our main contributions are three-fold:

$\bullet$ 
We propose a novel perspective of DG FAS that aligns feature representations on the fine-grained instance level instead of relying on artificially-defined domain labels.

$\bullet$ 
We present an innovative Instance-Aware Domain Generalization (IADG) framework, 
which actively simulates the instance-wise domain shifts
and whitens the style-sensitive feature correlation to improve the generalization.

$\bullet$  
Extensive experiments with analysis demonstrate the superiority of our method against state-of-the-art  competitors on the widely-used benchmark datasets.

\section{Related Work}
\subsection{Face Anti-Spoofing}
% \noindent \textbf{Face Anti-Spoofing.} 
% Face anti-spoofing (FAS) aims to recognize whether an image is taken from a real face or various face presentation attacks. 
Early studies used handcrafted features to tackle this problem, such as SIFT~\cite{2016Secure}, LBP~\cite{boulkenafet2015face,maatta2011face,LBP01}, and HOG~\cite{2014Context,HoG01}. Several works utilized HSV and YCrCb color spaces~\cite{boulkenafet2015face,boulkenafet2016face}, temporal information~\cite{siddiqui2016face,bao2009liveness}, and Fourier spectrum~\cite{li2004live} to address this issue. 
With the advent of CNN, some approaches model FAS with binary classification~\cite{DeepBinary00,DeepBinary01,DeepBinary02} or 
auxiliary supervision, \emph{e.g.,} depth map~\cite{yu2021revisiting}, reflection map~\cite{zhang2021structure} and r-ppg signal~\cite{lin2019face}. Recently, disentanglement~\cite{disentangle01,STCN} and custom operators~\cite{CDCN,BCN,chen2021dual} are also explored to improve the performance. 
Despite the gratifying progress in the intra-dataset settings, their performances still drop significantly on the target domains due to the large domain shifts across domains.
To address this issue, domain adaptation~\cite{zhou2020domain,zhou2022domain,zhou2023self,zhou2022context,PIT,zhou2020uncertainty} techniques have been recently introduced into FAS~\cite{zhou2022generative, wang2021self}, while the target data is not always accessible in real scenarios, which might fail these methods.
Hence, domain-generalizable FAS (DG-FAS) emerges, aiming to improve the performance on unseen domains. 
Based on adversarial learning~\cite{2018Domain,2019Multi,2020Single,wang2022domain} or meta-learning~\cite{liu2021dual,liu2021adaptive,2020Regularized,zhou2022adaptive,du2022energy} algorithms, almost all DG-FAS approaches rely on domain labels to learn domain-invariant representations. Nevertheless, such artificial domain labels are coarse-grained and subjective, which cannot reflect the real domain distributions accurately and may be unavailable due to huge labeling efforts. Besides, such domain-aware methods cannot guarantee that the learned representations are still insensitive to domain-specific styles. In this paper, we aim to address these issues from a new perspective by introducing instance-aware domain generalization to FAS.

\begin{figure*}
\includegraphics[width=1.0\textwidth]{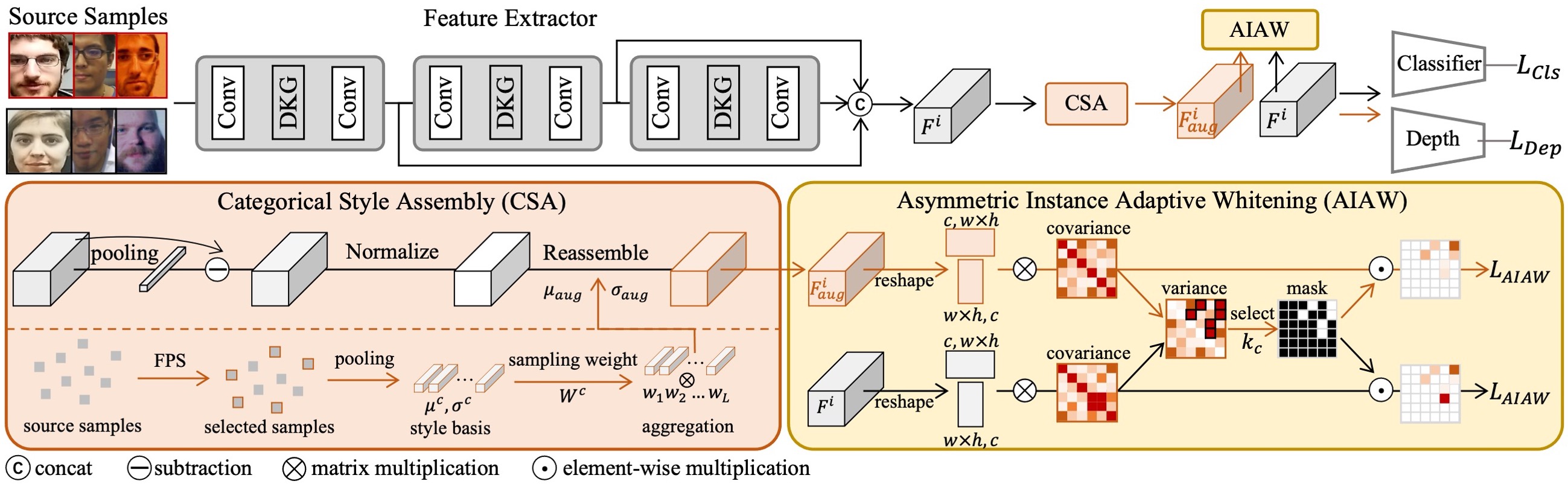}
% \vspace{-5mm}
\caption{Overview of the proposed  Instance-Aware Domain Generalization (IADG) framework for DG-FAS, mainly containing three key modules: Asymmetric Instance Adaptive Whitening(AIAW), Categorical Style Assembly (CSA), and Dynamic Kernel Generator (DKG). 
The whole framework actively generates style-diversified samples for each instance and adaptively eliminates the style-sensitive feature correlation, encouraging the instance-invariant feature to be insensitive to the domain-specific styles. Thus, we enhance the FAS model's generalizability towards unseen domains on a more fine-grained instance level without relying on any artificial domain labels. 
}
% \vspace{-3mm}
\label{fig:framework}
\end{figure*}

\subsection{Feature Covariance and Instance Whitening}
Previous works~\cite{gatys2015texture,gatys2016image} reveal that feature correlation (\textit{i.e.,} a covariance matrix) stores the domain-specific styles of images. In particular, the whitening transformation~\cite{li2017universal,pan2019switchable,cho2019image} aims to remove feature correlation and allows each feature to have unit variance.
Based on such theories, numerous studies have proved that whitening transformation is effective in removing the domain-specific styles in image translation~\cite{cho2019image}, style transfer~\cite{li2017universal}, and domain adaptation~\cite{roy2019unsupervised,sun2016deep}. Recent method~\cite{choi2021robustnet} applied the whitening loss for generalized semantic segmentation. Thus, instance whitening
may improve the generalization ability of features, but is still under-explored in DG FAS. Inspired by these works, considering the asymmetry between the real and spoof faces, we propose a novel Asymmetric Instance Adaptive Whitening to improve the generalization of FAS models. To the best of our knowledge,
this is the first work that reveals the potential of instance whitening for DG FAS.

\section{Methodology}
Figure~\ref{fig:framework} shows the overview of the proposed Instance-Aware Domain Generalization (IADG) framework, which aims to align the features on the instance level by weakening the features' sensitivity to instance-specific styles without the need for domain labels, improving the generalizability towards unseen domains. 
IADG includes three key components: Asymmetric Instance Adaptive Whitening (AIAW), Categorical Style Assembly (CSA), and Dynamic Kernel Generator (DKG). Firstly, AIAW aims to adaptively whiten the style-sensitive feature correlation for each instance from higher-order statistics on a fine-grained instance level.
Considering the discrepancies between real and spoof samples, AIAW also adopts asymmetric supervision on them to boost the generalization.
Furthermore, to facilitate the learning of style-insensitive features, CSA and DKG collaborate to generate style-diversified features for AIAW.
Specifically, DKG is designed to model the instance-specific features, which automatically generates instance-adaptive filters and works with static filters to facilitate comprehensive instance-aware feature learning. 
Based on such instance-specific features, CSA simulates instance-wise style shifts by considering the instance diversity to generate style-diversified samples in the wider feature space, which augments real and spoof faces separately to prevent the label changes in the FAS task. 
Next, we will describe each part of the method according to the order of model forwarding in the following sections.

\subsection{Dynamic Kernel Generator}
Considering the diversity of samples in multiple source domains, it is difficult to extract instance-adaptive features via one static filter.
Hence, we design DKG to automatically generate instance-adaptive filters, which assist the instance-static filter in learning the comprehensive instance-adaptive feature for further domain generalization.

\begin{figure}
\includegraphics[width=\linewidth]{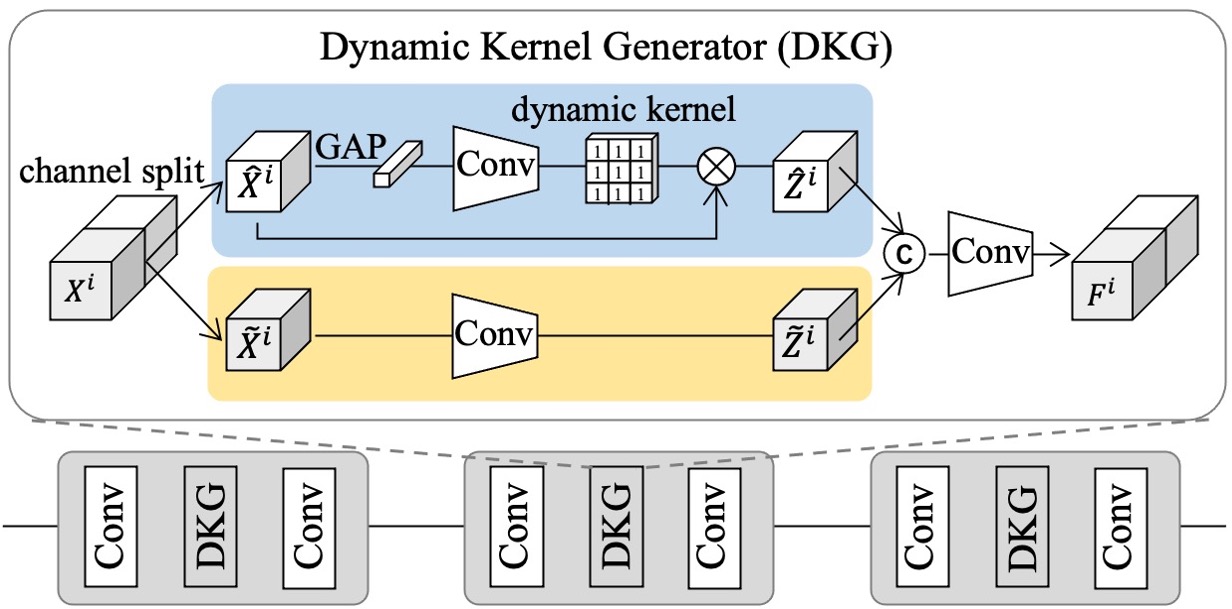}
% \vspace{-5mm}
\caption{Network structure of Dynamic Kernel Generator. 
}
% \vspace{-3mm}
\label{fig:DKG}
\end{figure}

DKG includes a static convolution branch and a dynamic kernel branch, where the former has constant parameters and the latter has parameters conditioned on each instance. The model is denoted static or dynamic depending on whether the model parameters vary with each sample.
As shown in Figure \ref{fig:DKG}, $X^i$ and $F^i$ are the input and output feature of the $i$-th sample of the DKG. Note that both two branches are jointly optimized during the training phase.
Specifically, we first split the channels of $X^i$ into two parts, which are denoted as $\hat{X}^i$ and $\Tilde{X}^i$.
In the bottom static convolution branch of Figure \ref{fig:DKG}, the feature $\tilde{X}^i$ with the latter half channels is forwarded into a static kernel $f_{\theta_{s}}$.
While in the top dynamic kernel branch, the feature $\hat{X}^i$ with the former half of the channels is forwarded into a global average pooling layer and a convolution block $f_{\theta_1}$, generating instance-adaptive kernels $W^{i}$.
Then, such instance-adaptive kernels $W^{i}$ are utilized to extract specific features of input features $\hat{X}^i$ via classical convolution.
We denote the output features of the static and dynamic branch as $\tilde{Z}^{i}$ and $\hat{Z}^{i}$,  respectively:
\begin{equation}
\label{eq:dynamic} 
\tilde{Z}_{i}=f_{\theta_{s}}\left(\tilde{X}_{i}\right), \hat{Z}^{i}=conv(\hat{X}^{i};W^{i}), 
\end{equation}
where $W^{i}=f_{\theta_{1}}(avgpool(\hat{X}^{i}))$ denotes the kernel parameters of dynamic convolution is dependent on the input instance $X^i$. 
Then, we concatenate $\tilde{Z}^{i}$ and $\hat{Z}^{i}$ at the channel dimension, and feed the results into a convolution block $f_{\theta_{2}}$ to output the feature, denoted as: 
\begin{equation}
\label{eq:output} 
F_{i}=f_{\theta_{2}} (concatenate(\tilde{Z}^{i}, \hat{Z}^{i}))
\end{equation}

\subsection{Categorical Style Assembly}
To simulate instance-wise style shifts in the wider feature space,  we propose Categorical Style Assembly (CSA) to generate style-diversified samples. 
Though AdaIN~\cite{huang2017arbitrary} in previous works shows the effectiveness in performing style augmentation,
these works just randomly swap or mix different source styles without considering the frequency of the source styles or the category information of the source samples. %it motivates us to improve such augmentation strategies in two aspects. 
Differently, we make two technical innovations here. 
Firstly, we design CSA by considering the diversity of various source samples to generate novel styles in a wider feature space.
Secondly, considering the specificity of
FAS task, we introduce the categorical concept into the CSA module and separately augment real and spoof samples to prevent the negative effects of label changes between different classes in the FAS task.

Now we describe CSA in detail. 
Inspired by farthest point sampling~(FPS)~\cite{qi2017pointnet++} that is widely used for point cloud down-sampling, we use FPS to select basis styles from all the source styles to ensure the style diversity. 
Specifically, we iteratively choose $L$ styles from all the samples for each class, \emph{e.g.,} real faces and spoof ones, such that the chosen samples have the most dissimilar styles with respect to the remaining samples. 
As a result, the basis styles obtained by FPS represent the whole style space to the utmost extent and also contain many rare styles that are far away from the dominant ones. 
Note that such basis styles are dynamically updated at every epoch (not fixed) since the style space is changing along with the model training. 
For the selected basis styles of each class, we calculate the mean $\mu$ and variance $\sigma^{2}$ and store them in memory banks.
Then, we use 
$\mu_{base}^{r}, \mu_{base}^{s}\in \mathbb{R}^{L\times L}$ to represent the mean of basis styles of real samples and spoof ones, respectively.

% $\mu^{real}_{base}$, $\mu^{spoof}_{base}$, $\sigma^{real}_{base}$ and  $\sigma^{spoof}_{base}$.

Moreover, considering that reassembling the content of real faces with the spoof styles might influence the liveness feature of real faces, we treat them differently in the feature augmentation. 
In other words, only when the content feature and the style feature have the same class label will we perform the style augmentation. 
Concretely, for each class $c$, we sample the combination weight $W^{c}=[w_1, \cdots, w_L]$ from Dirichlet distribution $B([\alpha_1, \cdots, \alpha_L])$, where the concentration parameters $[\alpha_1, \cdots, \alpha_L]$ are set to $1/L$ for $\alpha_l$. Then, the basis styles of each class $c$ are then linearly combined by $W^c$ for the aggregation: 
\begin{equation}
\begin{aligned}
     \mu^{c}_{aug} = W^{c}\cdot \mu^{c}_{base}, \qquad \sigma^{c}_{aug} = W^{c} \cdot \sigma^{c}_{base}, 
    \end{aligned}
\end{equation}
With the novel styles, style reassembled samples $F_{aug}$ are:
\begin{equation}
    F_{aug} = \sigma^{c}_{aug}\left(\frac{F_{org}-\mu(F_{org})}{\sigma(F_{org})}\right)+\mu^{c}_{aug},
\end{equation}
where statistics of new styles, \emph{i.e.,} mean $\mu_{aug}$ and standard deviation $\sigma_{aug}$, are chosen according to the class labels:
\vspace{2mm}
\begin{footnotesize}
\begin{equation}
\mu_{aug} = \left\{\begin{array}{l}
\! \mu_{aug}^{r}, \text { if } c=real\! \\
\! \mu_{aug}^{s}, \text {else} \end{array}\right.
\sigma_{aug} = \left\{\begin{array}{l}
\! \sigma_{aug}^{r}, \text { if } c=real\! \\
\! \sigma_{aug}^{s}, \text {else} \end{array}\right.
\vspace{2mm}
\end{equation}
\end{footnotesize}
As such, for each content feature of instance $F_{org}$, the new basis styles that have the same category as $F_{org}$ will be utilized for the style assembly, thus avoiding label changes and making the stylized samples more realistic.

\subsection{Asymmetric Instance Adaptive Whitening}

To align each sample in a finer granularity, we consider the correlation between the feature channels as explicit constraints for instance-adaptive generalization. 
Since previous studies have proved that feature covariance store domain-specific features and instance whitening is effective in removing such domain-specific styles in image translation~\cite{cho2019image}, style transfer~\cite{li2017universal}, and domain adaptation~\cite{roy2019unsupervised,sun2016deep}, it may improve the generalization ability of the features for DG-FAS. However, directly applying these instance whitening would inevitably remove the domain-invariant features that are discriminative for FAS classification, leading to less-desired performance. As such, it is non-trivial to design an innovative instance whitening loss for the FAS task. 

Specifically, we aim to selectively suppress the sensitive covariance and highlight the insensitive covariance.
Considering the context of the FAS task, we introduce the idea of asymmetry between the real and spoof faces into instance whitening: the real features should be more compact while the spoof features could be separated in the feature space. Thus, different selective ratios are applied to suppress sensitive covariance for real and spoof faces during the whitening. AIAW is computed as follows.
%Specifically, AIAW contains four steps.
Firstly, the feature map of a sample is fed into an instance normalization (IN) layer to output a normalized feature $F$. 
Then, the covariance matrix $\Sigma$ of $F$ is calculated as follows:
\begin{equation}
\Sigma=\frac{1}{H W}\left(F\right)\left(F\right)^{\top} \in \mathbb{R}^{C \times C}.
\end{equation}
Next, we derive the selective mask for the covariance matrix. 
Concretely, for the original features and style-augmented features, we calculate the variance $\sigma^{2}$ between the two covariance matrices $\Sigma_{org}$ and $\Sigma_{aug}$ as follows:
\begin{equation}
\begin{gathered}
\mu_{\Sigma}=\frac{1}{2}\left(\Sigma_{org}+\Sigma_{aug}\right), \\
\sigma_{\Sigma}^{2}=\frac{1}{2}\left(\left(\Sigma_{org}-\mu_{\Sigma}\right)^{2}+\left(\Sigma_{aug}-\mu_{\Sigma}\right)^{2}\right),\!
\end{gathered}
\end{equation}
where $\mu_{\Sigma}$ and $\sigma_{\Sigma}^{2}$ are respectively the mean and variance for each element from two different covariance matrices of the $i$-th image. 
We iterate each image and form a variance matrix: $V=\frac{1}{N} \sum \sigma_{\Sigma}^{2}$, where  $N$ is the number of images.
After that, we sort all variance elements of $V$ and select the top $k$ largest positions in the variance matrix $V$ to generate the Selective Mask ($M$), which is a binary classifier to 
distinguish which position is sensitive to the domain-specific 
styles. Compared to prior work~\cite{choi2021robustnet}, $M$ is more time-efficient during the training.
It can be formulated as: 
\begin{equation}
M_{i, j}(k_c)= \begin{cases}1, & \text { if } \mathrm{index}(V_{i, j}) < \mathrm{len}(V) \times k_c  \\ 0, & \text { otherwise }\end{cases}
\end{equation}
where the selective ratio  $k_c=k_r$ for real faces and $k_c=k_s$ for spoof ones.
Since the covariance matrix is symmetric, $M$ only
contains the strictly upper triangular part. Note that the selective ratio $k_r$ of real faces is larger than $k_s$ and then more sensitive covariance is suppressed, leading to more compact features. On the contrary, the spoof ratio $k_s$ is smaller and the extracted features are more dispersed.
% $k$ is set to $0.3\%$ empirically in all experiments. 

Finally, we adopt this mask to perform Asymmetric Instance Adaptive Whiting (AIAW), which pushes 
the selected covariance to 0.  The loss $\mathcal{L}_{\mathrm{AIAW}}$ aims to suppress the feature correlation in the selected positions as follows: 
\begin{equation}
\mathcal{L}_{\mathrm{AIAW}}= \sum_{k_c \in \{k_r, k_s \}} \sum_{t \in \{org, aug \}} \mathbb{E}\left[\left\|\Sigma_{t} \odot M(k_c) \right\|\right] ,
\end{equation}
where $\mathbb{E}$ is the the arithmetic mean. As such, the sensitive covariance will be suppressed, and the insensitive covariance will be highlighted. 
Moreover, different from prior works~~\cite{li2017universal,cho2019image,choi2021robustnet},
our mask is imposed on both the original and the augmented features, which means the proposed whitening is bilateral, to further guarantee that the suppressed covariance is still insensitive to domain-specific styles after style augmentation.

\begin{table*}[t!]
\centering
\begin{center}
\vspace{-1mm}
\resizebox{0.9\textwidth}{!}{%
\begin{tabular}{c | c c | c c | c c | c c }
\toprule
\multirow{2}{*}{\textbf{Methods}} &
\multicolumn{2}{c|}{\textbf{I\&C\&M to O}} &
\multicolumn{2}{c|}{\textbf{O\&C\&M to I}} &
\multicolumn{2}{c|}{\textbf{O\&C\&I to M}} &
\multicolumn{2}{c}{\textbf{O\&M\&I to C}} \\

&HTER(\%) &AUC(\%) &HTER(\%) &AUC(\%) &HTER(\%) &AUC(\%) &HTER(\%) &AUC(\%)\\
\midrule

LBPTOP~\cite{2014dynamic}&$53.15$ &$44.09$  &$49.45$ &$49.54$ &$36.90$ &$70.80$ &$42.60$ &$61.05$ \\

MS\_LBP~\cite{maatta2011face}&$50.29$ &$49.31$  &$50.30$ &$51.64$  &$29.76$ &$78.50$ &$54.28$ &$44.98$\\

ColorTexture~\cite{2017Face}&$63.59$ &$32.71$  &$40.40$ &$62.78$ &$28.09$ &$78.47$ &$30.58$ &$76.89$\\

Binary CNN~\cite{2014Learn}&$29.61$ &$77.54$  &$34.47$ &$65.88$  &$29.25$ &$82.87$ &$34.88$ &$71.94$\\

MMD-AAE~\cite{2018Domain}&$40.98$ &$63.08$  &$31.58$ &$75.18$ &$27.08$ &$83.19$  &$44.59$ &$58.29$\\

MADDG~\cite{2019Multi}&$27.98$ &$80.02$  &$22.19$ &$84.99$  &$17.69$ &$88.06$ &24.50 &84.51\\

RFM~\cite{2020Regularized}&16.45 &91.16  &17.30 &90.48 &13.89 &93.98 &20.27 &88.16\\

SSDG-M~\cite{2020Single}& 25.17 & 81.83 & 18.21 &\textbf{94.61}  &16.67 &90.47   & 23.11 & 85.45 \\

D$^2$AM~\cite{chen2021generalizable} &15.27 &90.87  &15.43 &91.22  &12.70 &95.66 &20.98 &85.58\\

DRDG~\cite{liu2021dual}&15.63 & 91.75  & 15.56 & 91.79 & 12.43 & 95.81  & 19.05 & 88.79\\

ANRL~\cite{liu2021adaptive}& 15.67 & 91.90   & 16.03 & 91.04 & 10.83 & 96.75  & 17.85 & 89.26\\
SSAN~\cite{wang2022domain} &19.51  &88.17  &14.00 &94.58  & 10.42 &94.76 & 16.47 &90.81 \\
AMEL~\cite{zhou2022adaptive} & 11.31 & 93.96  &18.60 &88.79 &10.23 &96.62 &11.88 &94.39\\
EBDG~\cite{du2022energy} & 15.66 & 92.02  &18.69 &92.28 &9.56 &97.17 &18.34 &90.01\\
\midrule
Ours (IADG) & \textbf{8.86} & \textbf{97.14}  &\textbf{10.62} &94.50 &\textbf{5.41} &\textbf{98.19} &\textbf{8.70} &\textbf{96.44}\\
\bottomrule
\end{tabular}}
\end{center}
\vspace{-5mm}
\caption{Comparison to the-state-of-art FAS methods on four testing domains. The bold numbers indicate the best performance. }
\label{tab:DG_SOTA_3to1}
\vspace{-2mm}
\end{table*}

\subsection{Overall Training and Optimization}
For high generalization capability on unseen domains, 
the classifier should capture consistent task-related information, even if there are perturbations in styles. Thus, both the original features $F_{org}^i$ and the augmented features $F_{aug}^i$ of the $i$-th input sample $X^i$ are used for supervision. 
% To ensure that the feature extractor extracts the task-related features $F^{i}$ for the $i$-th input sample $X^i$, 
Thus, we define a binary classification loss $\mathcal{L}_{\mathrm{Cls}}$ to ensure that the feature extractor extracts the task-related features:
\begin{small}
\begin{equation}
\begin{aligned}
    \mathcal{L}_{\mathrm{Cls}} =  -\sum_D Y^i_{cls} (log(Cls(F_{org}^{i})) + log(Cls(F_{aug}^{i})))
    % &-\sum_{(X_n,Y_n^{cls})}Y_{cls}^{n} log(Cls(F^{aug}_n))),
\end{aligned}
\end{equation}
\end{small}
where $Cls$ is the binary classifier detecting the face presentation attacks from the real ones, and $Y^i_{cls}$ is the classification label of $X^i$ sampled from domain $D$. 

\begin{table}[t!]
\centering
\begin{center}

\resizebox{0.48\textwidth}{!}{%
\begin{tabular}{c | c c | c c }
\toprule
\multirow{2}{*}{\textbf{Methods}} &
\multicolumn{2}{c|}{\textbf{M\&I to C}} &
\multicolumn{2}{c}{\textbf{M\&I to O}} \\
    &HTER(\%) &AUC(\%) &HTER(\%) &AUC(\%) \\
\midrule

MS\_LBP~\cite{maatta2011face} & $51.16$ & $52.09$ &  $43.63$ & $58.07$ \\
% IDA~\cite{2015Face} & $45.16$ & $58.80$ &  $54.52$ & $42.17$ \\
Color Texture~\cite{2017Face} & $55.17$ & $46.89$ &  $53.31$ & $45.16$ \\
LBPTOP~\cite{2014dynamic} & $45.27$ & $54.88$ &  $47.26$ & $50.21$ \\
% \midrule
MADDG~\cite{2019Multi} & $41.02$ & $64.33$ &  $39.35$ & $65.10$ \\
SSDG-M~\cite{2020Single} & $31.89$ & $71.29$ &  $36.01$ & $66.88$ \\
D$^2$AM~\cite{chen2021generalizable} &32.65 &72.04 &27.70 & 75.36 \\
DRDG~\cite{liu2021dual} &31.28 &71.50 &33.35 & 69.14 \\
ANRL~\cite{liu2021adaptive} & 31.06 & 72.12 & 30.73 &74.10 \\
SSAN~\cite{wang2022domain} & 30.00 & 76.20 & 29.44 & 76.62 \\
EBDG~\cite{du2022energy} & 27.97  & 75.84 & 25.94 & 78.28 \\
AMEL~\cite{zhou2022adaptive} &24.52 &82.12 & 19.68 & 87.01  \\
\midrule

Ours &\textbf{24.07} &\textbf{85.13} & \textbf{18.47} & \textbf{90.49}  \\
\bottomrule
\end{tabular}
}
\end{center}
\vspace{-5mm}
\caption{Comparison results on limited source domains. }
\label{tab:DG_SOTA_2to1}
\vspace{-3mm}
\end{table}

Since prior works~\cite{2018Learning,liu2021adaptive,liu2021dual} demonstrate that depth can be utilized as auxiliary information to supervise faces on the pixel level, we follow them using a depth estimator $Dep$, which estimates the depth maps for live faces and zero maps for spoof faces. %, to facilitate the learning of Feature Extractor. 
The depth loss $\mathcal{L}_{\mathrm{Dep}}$ is formulated as:

\begin{scriptsize}
\begin{equation}
% \begin{aligned}
    \mathcal{L}_{\mathrm{Dep}} = \sum_{D} \left\| Dep(F_{org}^i) - Y^i_{dep} \right\|_{2}^{2} + \left\| Dep(F_{aug}^i) - Y^i_{dep} \right\|_{2}^{2},
% \end{aligned}
\end{equation}
\end{scriptsize}
where $Y_{dep}^i$ denotes the depth label of the sample $X^i$. 

The total training loss $\mathcal{L}_{\mathrm{total}}$ is defined as:
\begin{equation}
\begin{aligned}
        \mathcal{L}_{\mathrm{total}} = \mathcal{L}_{\mathrm{Cls}} + \lambda  \mathcal{L}_{\mathrm{Dep}} +  \mathcal{L}_{\mathrm{AIAW}}.
\end{aligned}
\end{equation}
During training, the original branch and augmented branch are both optimized. During the inference phase, only the original branch is used.  $\mathcal{L}_{\mathrm{AIAW}}$ is used in training only.

\section{Experiments}
\subsection{Experimental Setting}
\label{sec:4.1}
\noindent \textbf{Datasets.} 
We use four public FAS datasets, \emph{i.e.,}  CASIA-MFSD \cite{Zhang2012A} (denoted as C), Idiap Replay-Attack \cite{2012Replay} (denoted as I), and MSU-MFSD \cite{2015Face} (denoted as M), OULU-NPU \cite{2017OULU} (denoted as O), to evaluate the effectiveness of our method. 
These four datasets are collected with various capture devices, different attack types, diverse illumination conditions, background scenes, and races. 
Thus, there exist large domain shifts among these datasets. 
In all experiments, we strictly follow the same protocols as previous DG FAS methods~\cite{2019Multi,2020Single,2020Regularized,liu2021dual,liu2021adaptive,zhou2022adaptive} for fair comparisons.

\noindent \textbf{Implementation details.} 
Our method is implemented with PyTorch and trained with Adam optimizer~\cite{kingma2014adam}. 
We use the same network architecture as ~\cite{liu2021dual,liu2021adaptive,2020Regularized}, and only the second convolutional block of each stage in the feature extractor is replaced by DKG.
We extract RGB channels of images, and the input size is $256\times256\times3$. 
For training, the hyper-parameter $\lambda$ is set to $0.1$, $L$ is set to 64, and $k_r =0.3\%$, $k_s =0.06\%$ empirically in all experiments. 
The learning rate is set to $0.0001$. 
Following prior works~\cite{liu2021adaptive,liu2021dual}, we utilize PRNet~\cite{2018Joint3D} to generate the pseudo-depth maps for depth supervision. 
The Half Total Error Rate (HTER) and the Area Under Curve (AUC) are used as evaluation  metrics. The code will be publicly available at this link: \href{https://github.com/qianyuzqy/IADG}{\textit{https://github.com/qianyuzqy/IADG}}.

\subsection{Comparisons to the State-of-the-art Methods}
\label{sec:4.2} Following common protocols~\cite{2019Multi,2020Single,liu2021dual,liu2021adaptive},  we perform Leave-One-Out (LOO) validation and domain generalization with limited source domains, respectively, to demonstrate the generalization towards the unseen domain. 

\noindent \textbf{Leave-One-Out (LOO).} As shown in Table~\ref{tab:DG_SOTA_3to1}, we conduct cross-domain generalization in four common Leave-One-Out (LOO) settings of the FAS task. Three datasets are randomly selected as source domains and the remaining one is treated as the unseen target domain, which is unavailable during the training process. 
The comparison methods in Table~\ref{tab:DG_SOTA_3to1} are divided into two parts: conventional FAS methods and DG FAS methods. 
From the table, we have the following observations. (1) Conventional FAS methods~\cite{2014dynamic,maatta2011face,2017Face,2014Learn} show unsatisfactory performances under these four cross-dataset benchmarks. This is because they do not consider learning generalizable  features across domains. (2) Our method outperforms most of these DG FAS methods~\cite{2018Domain,2019Multi,2020Single,2020Regularized,liu2021adaptive,liu2021dual,du2022energy,zhou2022adaptive} under four test settings. 
The main reason is that almost all of these methods heavily rely on artificially-defined domain labels, and the coarse-grained domain alignment cannot guarantee the extracted features are insensitive to domain-specific styles, leading to less-desired performances. In contrast to these domain-aware DG-FAS methods, we introduce instance-aware DG-FAS with significant improvements. 

% \vspace{-1mm}

\noindent \textbf{Limited source domains.}
As shown in Table~\ref{tab:DG_SOTA_2to1}, we validate our method when extremely limited source domains are available. Following prior works~\cite{liu2021dual,liu2021adaptive}, MSU-MFSD (M) and ReplayAttack (I) datasets are selected as the source domains for training, and the remaining two ones, \textit{i.e.}, CASIA-MFSD (C) and OULU-NPU (O), are respectively used as the target domains for testing. Our proposed method is superior to the state-of-the-art approaches by a large margin on the limited source data. This reveals that 
%no matter how many source domains exist, 
even in limited source domians, our instance-wise domain generalization is still effective toward unseen target domains, since our method does not require pulling all source domains together to perform domain alignment.

\subsection{Ablation Studies}
\label{sec:4.3} In this section, we first conduct ablation studies to study the contribution of each component.  
Then, we investigate the effect of different kernel designs, style augmentation strategies, and instance whitening losses. 
All ablation experiments are conducted on the I\&C\&M to O setting.
% All ablations are conducted on the I\&C\&M to O setting.

\noindent \textbf{Effectiveness of each component.} 
Table~\ref{table:ablation_component} shows the ablation studies of each component. The baseline means training the same backbone as~\cite{2019Multi,liu2021dual,liu2021adaptive,zhou2022adaptive} with IN layers, and the results are $19.75\%$ HTER and $87.46\%$ AUC. 
By adding DKG, we boost the performance to $16.94\%$ HTER and $90.14\%$ AUC. By further adding CSA, we effectively achieve $12.50\%$ HTER and $93.62\%$ AUC. 
Finally, our AIAW loss effectively lowers the HETR to $8.86\%$ and increases the AUC to $97.14\%$. 
These improvements confirm that these individual components are complementary and together they significantly promote the performance.
% These improvements confirm the effectiveness of individual components of our proposed approach and also reveals that these individual components are complementary and together they significantly promote the performance.

\begin{table}[t]
\centering

\resizebox{0.46\textwidth}{!}{%
\begin{tabular}{cccc|c|c} \toprule
 Baseline & DKG & CSA & AIAW & HTER(\%) & AUC(\%)\\
\midrule
 \checkmark  & - & -  &- & 19.75 & 87.46 \\
 \checkmark   & \checkmark  & - & - & 16.94 & 90.14 \\
\checkmark  & \checkmark  &  \checkmark & - & 12.50 & 93.62 \\
\checkmark  & \checkmark  &  \checkmark & \checkmark & 8.86 & 97.14 \\
\bottomrule
\end{tabular}
}
\vspace{-2mm}
\caption{Ablation of each component on I$\&$C$\&$M to O. }
\label{table:ablation_component}
\end{table}

\begin{table}[t]
\centering
% \resizebox{0.4\textwidth}{!}{%
\begin{tabular}{c|c|c} 
\toprule
Instance Whitening & HTER(\%) & AUC(\%)\\
\midrule
IW~\cite{li2017universal} & 14.89 & 91.51 \\
GIW~\cite{cho2019image} & 14.13 & 92.68 \\
ISW~\cite{choi2021robustnet} & 11.97 & 94.25 \\
\midrule
Ours (AIAW) & 8.86 & 97.14 \\
\bottomrule
\end{tabular}
% }
\vspace{-2mm}
\caption{Ablation of instance whitening on I$\&$C$\&$M to O. }
\label{table:ablation_whitening}
\end{table}

\begin{table}[t]
\centering
\resizebox{0.40\textwidth}{!}{%
\begin{tabular}{c|c|c|c} 
\toprule
Instance Whitening &$k_{r}$:$k_{s}$& HTER(\%) & AUC(\%)\\
\midrule
Symmetric IAW &1:1 & 10.69 & 95.70 \\
\midrule
\multirow{4}{*}{\text { Asymmetric IAW }} 
&1:0.8 & 10.24 & 96.10 \\
&1:0.5 & 10.20 & 96.37 \\
&1:0.2 & \textbf{8.86} & \textbf{97.14} \\
&1:0.1 & 9.86 & 96.62 \\
\bottomrule
\end{tabular}
}
\vspace{-2mm}
\caption{Effect of instance whitening on I$\&$C$\&$M to O. }
\label{table:ablation_whitening2}
\vspace{-2mm}
\end{table}

\noindent \textbf{Comparisons of different Instance Whitening losses.} As illustrated in Table~\ref{table:ablation_whitening}, we show the comparison of different whitening losses.  IW~\cite{li2017universal} and GIW~\cite{cho2019image} suppress all covariance elements in the upper triangular of the covariance matrix, which would inevitably eliminate the domain-invariant features that are discriminative for classification, thus affecting the performance. ISW~\cite{choi2021robustnet} shows better results. However, all these losses do not consider the asymmetry between the real and spoof faces, which is vital for the FAS task. Besides, all of them only constraining the covariance of the original feature cannot guarantee that the suppressed covariance is still insensitive to domain-specific styles after style augmentation. In contrast, our AIAW loss is asymmetric and bilateral and shows superior results.

\noindent \textbf{Effect of various Instance Adaptive Whitening losses.} Table~\ref{table:ablation_whitening2} shows the effect of different IAW losses. Symmetric IAW means the selective ratio $k_r$:$k_s$=1:1, and real and spoof faces are equally whitened. We observe that Asymmetric IAW is superior to Symmetric IAW, indicating the necessity of more strict constraints on real people. 
This is because spoof faces tend to have larger variance than real ones, and it does not make sense to use too much constraint to force alignment between different kinds of attacks.
And the highest performance is reached when $k_r$:$k_s$=1:0.2, indicating that the suppression of spoof covariance cannot be too large. And constraining spoof faces can also provide a certain promotion effect even though the constraint is small.

% \vspace{-3mm}
\begin{table}[t]
\centering
% \resizebox{0.4\textwidth}{!}{%
\begin{tabular}{c|c|c} 
\toprule
Style Augmentation & HTER(\%) & AUC(\%)\\
\midrule
MixStyle~\cite{zhou2021mixstyle} &15.00 &92.72 \\
SSA~\cite{wang2022domain} & 13.29 & 93.38 \\
SHM~\cite{zhao2022style} & 11.18 & 94.34 \\
\midrule
% Ours (CSA) & 10.69 & 95.70 \\
Ours (CSA) & 8.86 & 97.14 \\
\bottomrule
\end{tabular}
% }
\vspace{-2mm}
\caption{Ablation of style augmentation on I$\&$C$\&$M to O. }
\label{table:ablation_augmentation}
% \vspace{-3mm}
\end{table}

% \vspace{-2mm}
\begin{table}[t]
\vspace{-2mm}
\centering
% \resizebox{0.4\textwidth}{!}{%
\begin{tabular}{c|c|c} 
\toprule
Kernel Designs & HTER(\%) & AUC(\%)\\
\midrule
Static Conv Only & 13.33 & 92.81 \\
Dynamic Conv Only & 12.18 & 94.37 \\
\midrule
Ours (DKG) & 8.6 & 97.14 \\
\bottomrule
\end{tabular}
% }
\vspace{-2mm}
\caption{Ablation of DKG designs on I$\&$C$\&$M to O. }
\label{table:ablation_kernel}
% \vspace{-3mm}
\end{table}

\begin{figure}[t!]
\vspace{-1mm}
\centering
\includegraphics[width=0.47\textwidth]{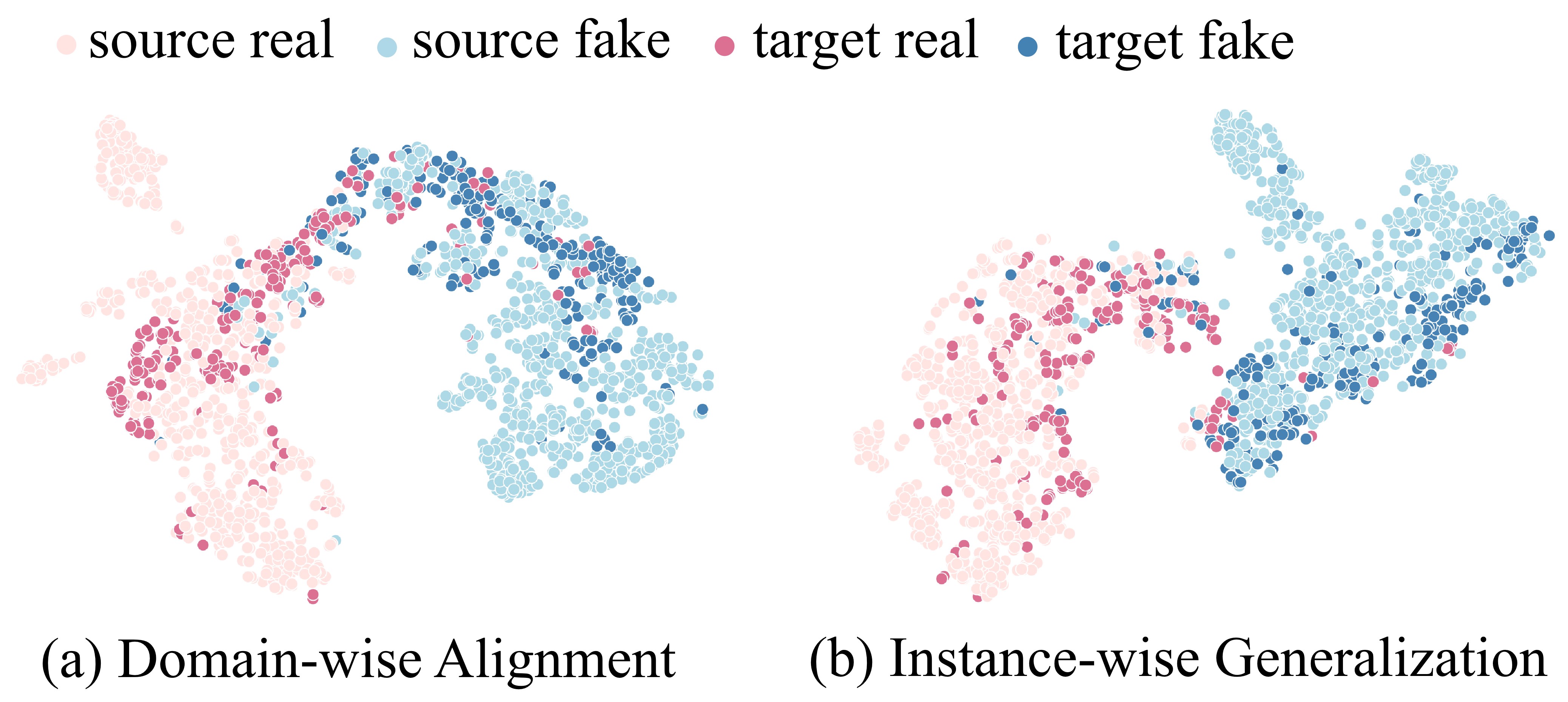}
% \vspace{-7mm}
\vspace{-2mm}
\caption{Comparison results of t-SNE feature visualization. }
\label{fig:tsne_comparison}
\vspace{-5mm}
\end{figure}

\noindent \textbf{Comparisons of different style augmentations.} 
Table~\ref{table:ablation_augmentation} shows the effect of different style augmentations. Mixstyle~\cite{zhou2021mixstyle} and SSA~\cite{wang2022domain} only yield limited improvements. SHM~\cite{zhao2022style} achieves a better performance; however, it is still inferior to our method. The main reasons lie in two aspects: 1) directly mixing styles tends to generate more samples of dominant styles, and the generated distributions may still involve a huge discrepancy from the real-world scenarios. In contrast, our method generates novel styles by increasing style diversity. 2) They might change the label when reassembling features with different classes. 
Compared to them, our CSA introduces the category concept, preventing label changes and unrealistic feature generation.

\noindent \textbf{Ablations of different DKG designs.}
Table~\ref{table:ablation_kernel} illustrates the ablations of different DKG designs. Firstly, we find that by replacing our DKG module with a static convolution branch degrades the performances and only achieves $13.33\%$ HTER and $92.81\%$ AUC.
Besides, merely using the dynamic kernel achieves $12.18\%$ HTER and $94.37\%$ AUC. This reveals that only a static or dynamic kernel is not sufficient in adapting to various unseen domains.
Compared to them, we find their combination is the best choice. This is because the dynamic kernel automatically generates instance-wise filters to assist comprehensive instance-specific
feature learning when combined with static filters.

% This is because only making the model dynamic cannot guarantee the learning of the instance-invariant feature. 
% Compared to them, we find the combination of the dynamic branch and static branch is the best choice, since we can learn the instance-invariant feature via the static kernel and then adapt it to each instance, thus learning instance-specific features. 

\begin{figure}[t!]
\centering
\includegraphics[width=\linewidth]{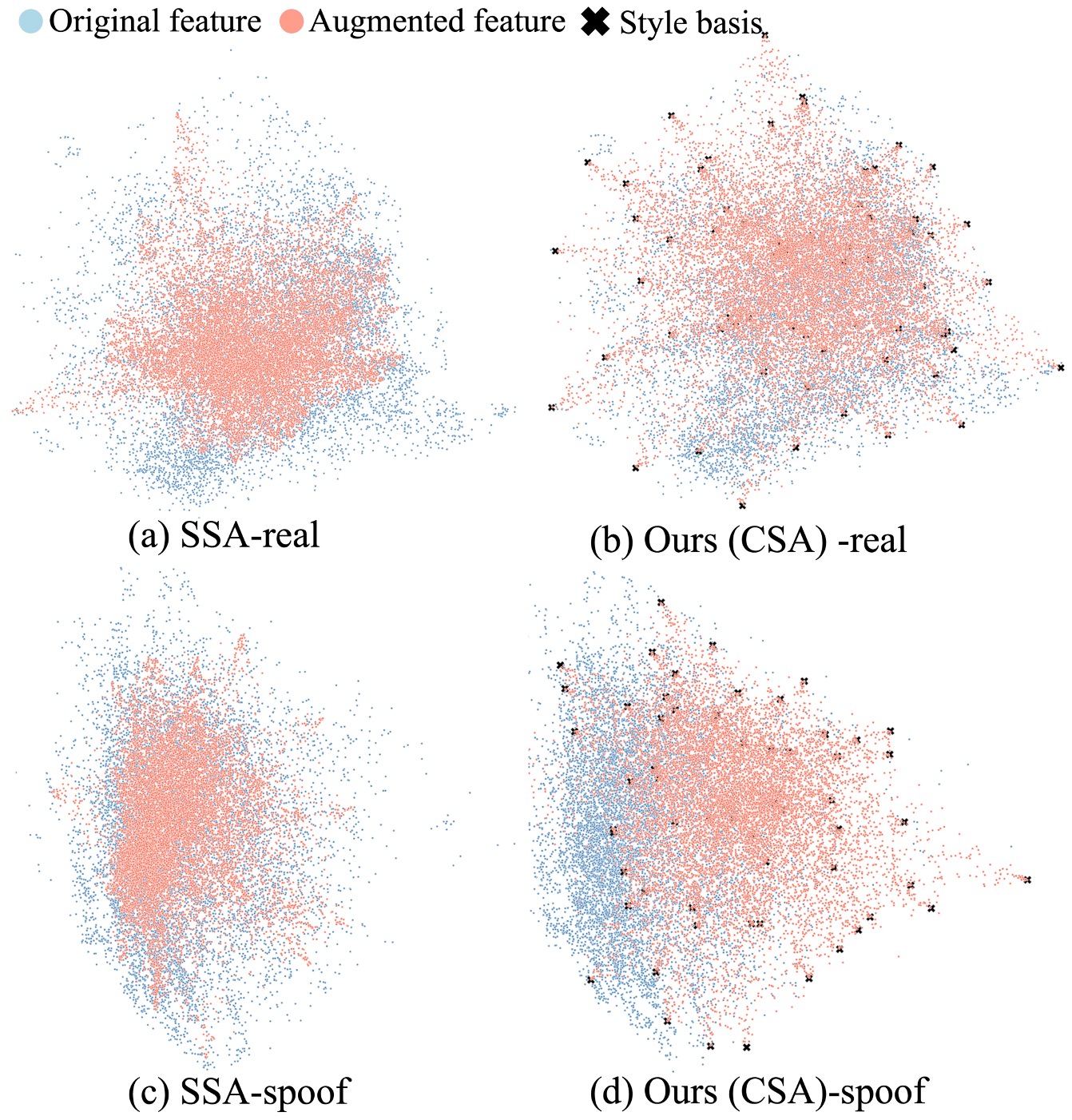}
% \vspace{-4mm}
\caption{T-SNE feature visualization of different style augmentations. CSA generates more diversified features than SSA~\cite{wang2022domain}. }
\label{fig:tsne_aug}
\vspace{-2mm}
\end{figure}

\subsection{Visualization and Analysis}
\label{sec:4.4}
\noindent \textbf{T-SNE visualization of feature distributions.} 
To understand how the IADG framework aligns the feature, we utilize 
t-SNE to visualize the feature distributions of each domain. 
From the figure, we can make the following two observations: 1) As shown in Figure~\ref{fig:tsne_comparison}~(a), previous method that performs the domain-aware alignment could well discriminate the source data by binary classification.
However, the target data is not well classified near the decision boundary. Instead, in Figure~\ref{fig:tsne_comparison}~(b) our approach manages to learn a better decision boundary between the real and fake samples. 2) By performing instance-aware domain generalization, the real features are relatively more compact, while the distributions of domain-wise alignment are much looser. The reason is that by breaking down source domain barriers, the alignment of various source domains is significantly simplified, and learning domain-invariant features is statistically achieved by generalizing the model per instance. 

\noindent \textbf{T-SNE visualization of syle basis and augmented features.} 
Figure~\ref{fig:tsne_aug} visualizes the t-SNE distributions of the style basis and augmented features. We have the following two observations: 1) The style basis selected by the FPS cover almost all possible source samples. %to the utmost extent. 
2) When using SSA~\cite{wang2022domain} for style augmentation, the generated features are still within the source distributions and even ignore some possible rare styles. In contrast, our method generates more diverse styles and even generates some novel styles of out-of-the-source distributions, especially for spoof samples, which enhances the generalization ability.

\begin{figure}[t!]
\centering
\includegraphics[width=0.49\textwidth]{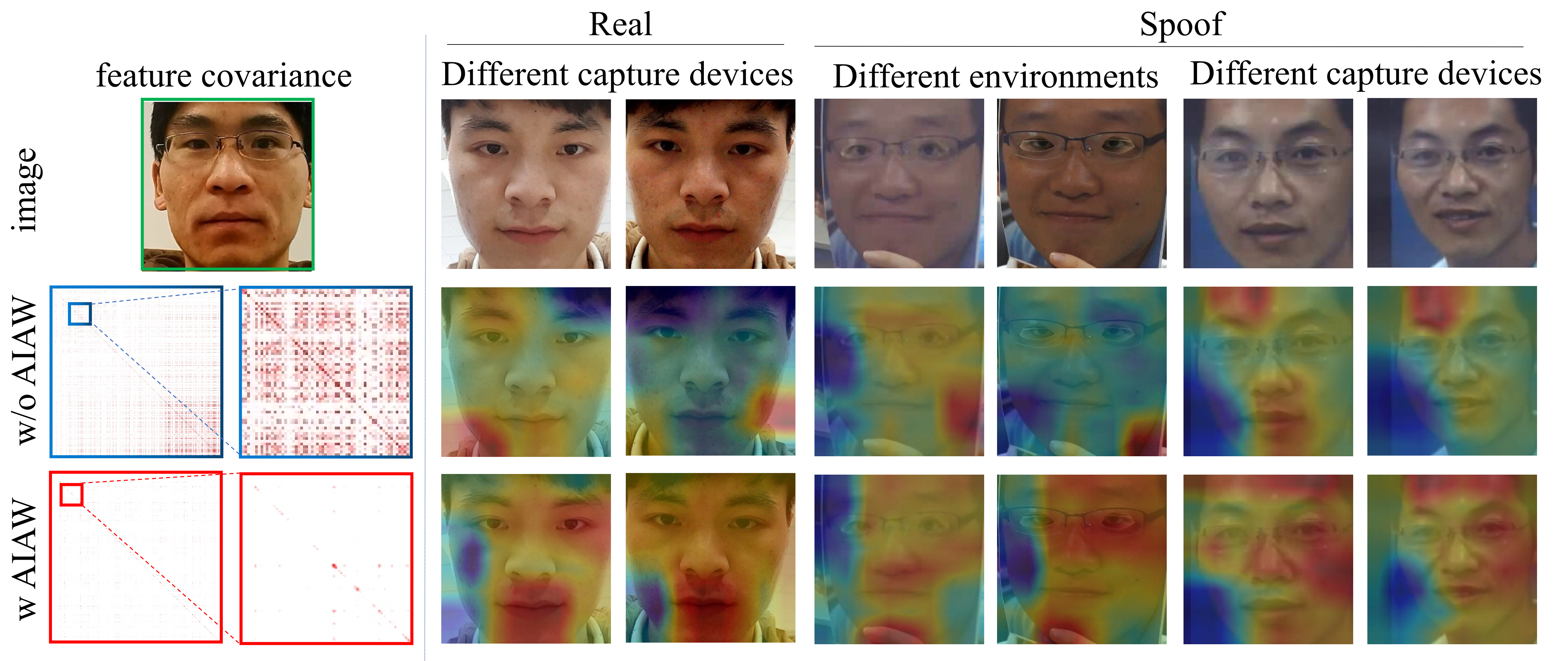}
% \vspace{-5mm}
\vspace{-5mm}
\caption{Visualization of covariance and Grad-CAM~\cite{zhou2016learning} activation maps under different capture devices and environments. }
\vspace{-3mm}
\label{fig:vis_kernel}
\end{figure}

\noindent \textbf{Visualization of feature covariance and Grad-CAM feature visualizations.} To illustrate how AIAW works, we visualize the feature covariance and Grad-CAM~\cite{zhou2016learning} activation maps. As shown in Figure \ref{fig:vis_kernel}, compared to our method without using AIAW (the second row), most areas of feature covariance are whitened but a small number of covariance elements remain large, showing our AIAW (the last row) selectively eliminates the covariance. Besides, if AIAW is not used (the second row), the attention shifts in different scenes. In contrast, our method (the last row) concentrates more on faces under different capture devices and environments, which demonstrates the effectiveness of AIAW.

\section{Conclusion}
In this paper, we propose a novel perspective of DG FAS that aligns features on the instance level without the need for domain labels. 
Concretely, we present a new Instance-Aware Domain Generalization (IADG) framework to learn the generalizable feature by weakening the features' sensitivity to instance-specific styles.
Specifically, we first propose Asymmetric Instance Adaptive Whitening to adaptively eliminate the 
style-sensitive feature correlation for each instance by considering the asymmetry between live and spoof faces, improving the generalization. Then, Dynamic Kernel Generator and Categorical Style Assembly  are proposed to first extract the instance-specific features and then generate the style-diversified features with large style shifts, respectively, further facilitating the learning of style-insensitive features. Extensive experiments and analysis on several benchmark datasets demonstrate the superiority  of our method over state-of-the-art competitors. 

\section{Acknowledgment}
This work was supported by the National Natural Science Foundation of China (72192821, 61972157), Shanghai Municipal Science and Technology Major Project (2021SHZDZX0102), Shanghai Science and Technology Commission (21511101200), Shanghai Sailing Program (22YF1420300, 23YF1410500), CCF-Tencent Open Research Fund (RAGR20220121) and Young Elite Scientists Sponsorship Program by CAST (2022QNRC001).

%%%%%%%%% REFERENCES
{\small
\bibliographystyle{ieee_fullname}
\bibliography{egbib}
}

\end{document}